\documentclass[runningheads]{llncs}
\usepackage{graphicx}
\usepackage[english]{babel}
\usepackage[utf8]{inputenc}
\usepackage{algorithm}
\usepackage{arevmath}     
\usepackage[noend]{algpseudocode}
\usepackage{amsmath}
\usepackage{booktabs,dcolumn}
\usepackage{hyperref}
\begin{document}
\title{Regularized Weight Aggregation in Networked Federated Learning for Glioblastoma Segmentation}
\titlerunning{Federated Regularized Weight Aggregation}
%
\author{Muhammad Irfan Khan\inst{1} \and
Mohammad Ayyaz Azeem\inst{2} \and
Esa Alhoniemi\inst{1} \and
Elina Kontio\inst{1} \and
Suleiman A. Khan\inst{1} \and
Mojtaba Jafaritadi\inst{1}}
\authorrunning{Khan et al., 2022}
%
\institute{Turku University of Applied Sciences, Turku 20520, Finland \and
Riphah International University, Islamabad 45210, Pakistan 
\\
\email{irfan.khan@turkuamk.fi,32175@student.riphah.edu.pk, mojtaba.jafaritadi,esa.alhoniemi,elina.kontio, suleiman.alikhan@turkuamk.fi}
} 
\maketitle              

\begin{abstract}

In federated learning (FL), the global model at the server requires an efficient mechanism for weight aggregation and a systematic strategy for collaboration selection to manage and optimize communication payload. We introduce a practical and cost-efficient method for regularized weight aggregation and  propose a laborsaving technique to select collaborators per round. We illustrate the performance of our method, regularized similarity weight aggregation (RegSimAgg), on the Federated Tumor Segmentation (FeTS) 2022 challenge's federated training (weight aggregation) problem. Our scalable approach is principled, frugal, and suitable for heterogeneous non-IID collaborators. Using FeTS2021 evaluation criterion, our proposed algorithm RegSimAgg stands at 3rd position in the final rankings of FeTS2022 challenge in the weight aggregation task. Our solution is open sourced at: \url{https://github.com/dskhanirfan/FeTS2022}

\end{abstract}

\section{Introduction}

Federated learning (FL) is on the horizon to replace the current paradigm of data sharing, allowing for privacy-preserving cross-institutional research including a wide range of biomedical disciplines. In simple terms, FL is a machine learning paradigm in a distributed or decentralized setting. 
It is particularly favorable because pooling all the curated data from different data silos to a central location is arduous, for training machine learning models. Moreover, sharing sensitive data is becoming increasingly difficult due to data privacy and security concerns, bureaucratic challenges and stringent GDPR (EU) and HIPAA (US) laws~\cite{annas2003hipaa,voigt2017eu}.

\begin{figure*}[!th]
\centering\includegraphics[width=0.9 \textwidth]{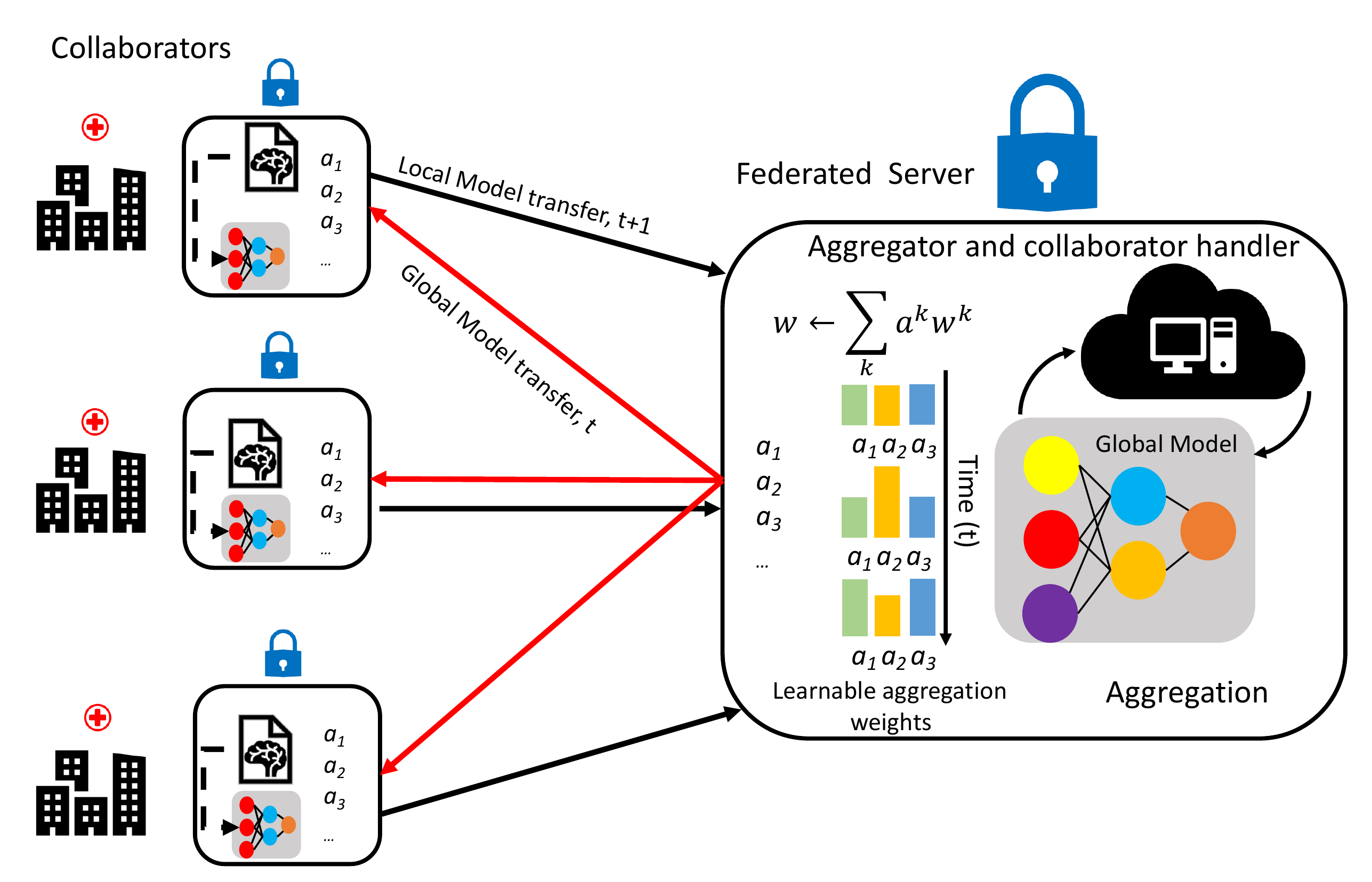}
\caption{General workflow of an FL-trained model and the key components in a federated learning setting~\cite{xu2021federated}. Private collaborators communicate the local weight updates with a federated secure server at regularly occurring intervals to learn a global model; the server aggregates the updates and sends back the parameters of the updated global model to the clients. The aggregation weights are learned on the clients and dynamically adjusted throughout the
training process when communicating with the federated server.}
\label{FL}
\end{figure*}

The implementation of FL in practice requires several distinct clients called \textit{collaborators} to contribute to the creation of a global expert model via a defined server called an \textit{aggregator}. Each collaborator provides some of the information that the aggregator would combine~\cite{mcmahan2017communication}. 
The aggregator does not have access to the collaborator's private data, which does not egress from the collaborator. The actual training tasks are executed at the collaborators, because each collaborator has a chunk of the whole data. During federation rounds, participating collaborators cartel the parameters combined at the server for a unified consensus model to foster knowledge exchange. This global model is assembled by learned information, in terms of parameters, from a conglomerate of individual participating collaborators that execute the learning task independently. Hence, model training is essentially performed in a distributed fashion on the data hoarded at distinct edge devices. After model training, the generalizable global model or model parameters are dispatched to the collaborators~\cite{Kairouz2019}. Consequently, in a federation process, all the participating collaborators get potential benefit from the global model at the aggregator, because they receive the learned model parameters collected from trained models on other collaborators as well. Moreover, the performance and inference of global model, trained on several collaborators, on unseen data is better as compared to the model trained on any individual collaborator data. Figure~\ref{FL} shows a high-level schema of the federated learning framework.

The utility spectrum of federated learning is quite broad with wide-range applications in telecommunication services, fintech, and healthcare sector. In healthcare, data analytics in radiology and genomics can particularly welfare from harnessing the power of FL ecosystem. A direct clinical impact of FL therefore could be made in the most-effective translation medicine, e.g.\ privacy-preserving medical image segmentation \cite{li2019privacy,sarma2021federated}. However, training a shared generalizable global model for medical image segmentation in a federated fashion presents a number of challenges and limitations. These include training local models using different imaging modalities on many scanners from different manufacturers, acquisition times, and various image resolutions and protocols. The improper use of such training data, referred to as not independent and identically distributed (non-IID), thus can result in performance degradation among different clients. FL has been combined with domain adaptation \cite{yan2020variation}, contrastive learning \cite{li2021model}, and knowledge distillation \cite{liu2021feddg} in order to learn a more generalizable federated model. Other limitations include cross domain and imbalance of annotated data (limited labeling budgets) \cite{bernecker2022fednorm}. The challenge of data heterogeneity and domain shifting was recently tackled in novel ways by, for example, federated disentanglement learning via disentangling the parameter space into shape and appearance \cite{bernecker2022fednorm} and automated federated averaging based on Dirichlet distribution \cite{xia2021auto}. Dynamic Re-Weighting mechanisms \cite{liu2022ms}, federated cross ensemble learning \cite{xu2022federated}, and label-agnostic (mixed labels) unified FL formed by a mixture of the client distributions \cite{wicaksana2022fedmix} have been recently proposed to relax an unrealistic assumption that each client's training set will be annotated similarly and therefore follows the same image supervision level during the training of an image segmentation model. Although extensive research has been carried out on FL, there is still a need for methods to enable the development of more generalized FL models for clinical use which can effectively deal with statistical heterogeneity in weight aggregation, communication efficiency, and privacy with security.

In this paper, we aim to establish an adaptive regularized weight aggregation by upgrading our previously developed similarity weight aggregation (SimAgg) algorithm~\cite{KhanSimAgg}. We propose a robust and efficient federated lesion segmentation algorithm applicable in generalized and realistic detection of the “rare” disease of glioblastoma, a form of brain cancerous tumor, and particularly on the delineation of its sub-regions by leveraging multi-modal magnetic resonance imaging (MRI) brain scans~\cite{pati2021federated}. We present an extensive evaluation of the proposed regularized similarity weight aggregation (RegSimAgg) strategy in a networked federated learning fashion. In the light of our previous research in FeTS2021, we propose an efficient yet simple method for addressing FeTS2022 and existing challenges of FL.

The rest of this paper is organized as follows: in Section~\ref{sec:method}, we describe the upgraded methodologies to the previous SimAgg algorithm including collaborator selection and weight aggregation regularization through our experiment setting. In Section~\ref{sec:results}, we describe FL experiments and evaluate the performance of the proposed method quantitatively and in Section~\ref{sec:discussion}, we discuss about the presented work, potentials and limitations, and describe our future direction in FL. Finally, Section~\ref{sec:conclusion} concludes this work.

\section{Methods}
\label{sec:method}

\subsection{FeTS 2022 challenge}
Federated Tumor Segmentation (FeTS) 2022 is a continuation to the previous FeTS2021 \cite{pati2021federated} challenge with the focus on federated training methodologies including weight aggregation, client selection, training per-round, compression, communication efficiency, and algorithmic generalizability on out-of-sample data. It is intended to build and evaluate a consensus model that effectively identifies intrinsically heterogeneous brain tumors. 

The FeTS 2022 challenge provides updated multi-institutional multi-parametric Magnetic Resonance Imaging (mpMRI) scans of glioblastoma (GBM), the most common primary brain tumor, prior to any kind of resection surgery. The datasets used in the FeTS 2022 challenge are the subset of GBM cases from the Brain Tumor Segmentation (BraTS) Continuous Challenge which aims at identifying state-of-the-art segmentation algorithms for brain diffuse glioma patients and their sub-regions~\cite{baid2021rsna,bakas2017segmentationGBM,bakas2017segmentationLGG,bakas2017advancing,menze2014multimodal}. 

All FeTS brain MRI scans, provided as NIfTI files (\texttt{.nii.gz}), had four structural MRI sequences including native (T1), post-contrast T1-weighted (T1Gd), T2-weighted (T2), and T2 FLuid Attenuated Inversion Recovery (FLAIR) volumes. Image samples were acquired with different clinical protocols and various scanners from multiple data contributing institutions. One to four raters annotated each of the images manually, following a standardized protocol, and their annotations were approved by neuroradiologists. Annotations comprise the pathologically confirmed segmentation labels with similar volume size of 240$\times$ 240$\times$155 including the GD-enhancing tumor (ET — label 4), the peritumoral edematous/invaded tissue (ED — label 2), and the necrotic tumor core (NCR — label 1). All these provided MRI scans including the ground truth data were pre-processed such as rigid registration, brain extraction, alignment, 1$\times$1$\times$1 mm resolution resampling, and skull stripping were applied as described in~\cite{menze2014multimodal}.

The training and validation datasets include 1251 and 219 subjects, respectively. The training set consists of two partitions each providing information for how to split the training data into non-IID institutional subsets. That is, each patient dataset is linked to a de-identified partitioning label according to the acquiring institutions.


We deployed Intel Federated Learning (OpenFL)~\cite{reina2021openfl} framework for training brain tumor segmentation model --- an encoder-decoder 3D U-shape type of convolutional neural network provided by FeTS2022 challenge --- using the data-private collaborative learning paradigm of FL \cite{pati2022federated}. OpenFL considers two main components: 1) the collaborator which uses a local dataset to train the global model and 2) the aggregator which receives model updates from each collaborator and fuses them to form the global model. 

\subsection{Regularized Similarity Weighted Aggregation (RegSimAgg)} \label{m1}
In FeTS 2021 challenge, we suggested a novel aggregation method named Similarity Weighted Aggregation (SimAgg) for efficient aggregation of model parameters at the server that is suitable for both IID as well as non-IID data~\cite{KhanSimAgg}. Here, we propose an extension of SimAgg method which contains a regularization mechanism to speed up convergence, which is a critical issue in the computationally demanding federated learning framework.

\subsubsection{Collaborator Selection.}
\begin{figure*}[!th]
\centering\includegraphics[width=1\textwidth]{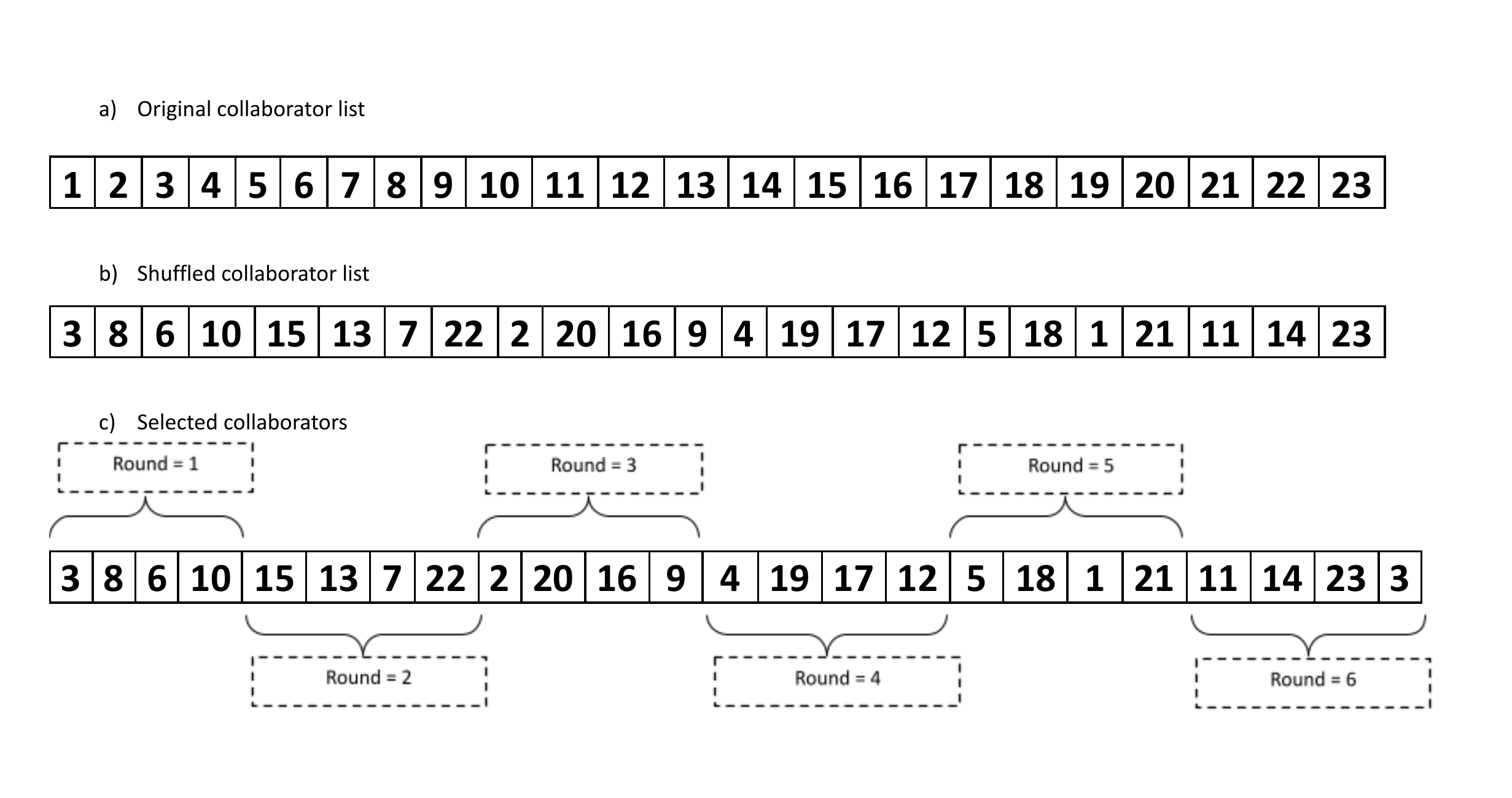}
\caption{Strategy for choosing collaborators. In a) Original collaborator list is given to the model, in b) collaborator order is randomized, and in c) collaborators are selected for each round using a sliding window. The collaborator list is reshuffled once it has been fully utilized, and the process repeats step b). The collaborators chosen in a certain combination during one FL round, however, will not be chosen again during subsequent rounds.
}
\label{collaborators}
\end{figure*}

 We allow collaborators to contribute in a nondeterministic fashion by picking up a subset of the available collaborators (for example, 20\%) at each round. To ensure that the model sees all collaborators the same number of times at regular intervals, we use a sliding window over the randomized collaborator index as shown in Fig.~\ref{collaborators}; once all collaborators have participated in the updates, the order is randomly shuffled. In this manner, we ensure roughly equal participation of all collaborators. However, a particular combination of the collaborators selected in one FL round will not be repeated in the successive rounds.
 

\subsubsection{Weight Aggregation.}
The fact that model parameters from the collaborators can differ is a major concern with non-IID data. We employ weighted aggregation of the collaborators at the server to overcome such a scenario. The collaborators' weights are determined by measuring similarity of the collaborators to their non-weighted average. We have also learned that, from the convergence point of view, it is beneficial to add regularization to the weighting process after the initial FL rounds. Our aggregation algorithm is compactly described in Algorithm~\ref{simagg_algo}, and the steps are explained in detail below.

At round $r$, the parameters $p_{C^r}$ of the participating collaborators $C^r$ are sent to the server. At the server, the average of these parameters is calculated as:
\begin{equation}
\hat{p} = \frac{1}{\lvert C^r \lvert}\Sigma_{i\in C^r}{p_i}.
\label{eqn_pavg}
\end{equation}
We then calculate the inverse distance (similarity) of each collaborator $c \in C^r$ from the average: 
\begin{equation}
sim_c = \frac{\Sigma_{i \in C^r}{\lvert p_i - \hat{p} \rvert}}{\lvert p_c - \hat{p} \rvert + \epsilon},
\label{eqn_sim}
\end{equation}
where $\epsilon = 1e-5$ (small positive constant). We normalize the distances to obtain {\em similarity weights} as follows:

\begin{equation}
u_c = \frac{sim_c}{\Sigma_{i\in C^r}{sim_i}}.
\label{eqn_sim_weights}
\end{equation}

The collaborators close to the average receive a high similarity weight and vice versa. In the extreme case this approach can expel the diverging collaborator.

In order to adjust for the effect of varying number of samples at each collaborator $c \in C^r$, we use   {\em sample size weights} that favor collaborators with large sample sizes:

\begin{equation}
v_c = \frac{N_c}{\Sigma_{i\in C^r}{N_i}},
\label{eqn_sample_weights}
\end{equation}
where $N_c$ is the number of examples at collaborator $c$. 

Using the weights obtained using Eqs.~\ref{eqn_sim_weights} and~\ref{eqn_sample_weights}, the {\em aggregation weights} are computed as:

\begin{equation}
w_c = \frac{u_c+v_c}{\Sigma_{i\in C^r}{(u_i+v_i)}},
\label{eqn_fsim_weights}
\end{equation}

If we have run enough iterations (e.g., $r > 10$), we also regularize the aggregation weights:

\begin{equation}
w_c = \frac{w_c}{\frac{1}{|C^r|}\sum_{i\in C^r} \left( p^{\rm prev}_i - p_i \right)}.
\label{eqn_fsim_regularization}
\end{equation}

Compared to the SimAgg method proposed earlier by us in \cite{KhanSimAgg}, this is the only but remarkable change in our algorithm. The basic idea is to let the FL system learn fast during the initial FL rounds after which the regularization makes learning somewhat slower but stable by suppressing significant weight adjustments. The FL round iteration limit (here 10) for starting the regularization is a hyperparameter of the FL aggregator, and based on our experience it is sensible to set it to a relatively small value.

The parameters are finally aggregated as a weighted average using the aggregation weights:
\begin{equation}
p^m = \frac{1}{\lvert C^r \lvert} \cdot \Sigma_{i\in C^r}{(w_i \cdot p_i)}.
\label{eqn_params}
\end{equation}

The normalized aggregated parameters $p^m$ are dispatched to the next set of collaborators in the successive federation rounds.

\begin{algorithm}
\caption{RegSimAgg aggregation algorithm}
\label{simagg_algo}
\begin{algorithmic}[1]
\Procedure{Regularized Similarity Weighted Aggregation}{$C^r$, $p_{C^r}$, $p_{C^r}^{\rm prev}$}

    \State $\epsilon$ $\leftarrow$ $1e-5$ \Comment{$C^r$ = set of collaborators (at round $r$)}
    
    \State  $\hat{p}$ = average($p_{C^r}$) using \textbf{Eq.~\ref{eqn_pavg}} \Comment{$p_{C^r}$ = parameters of the collaborators in $C^r$}
    
    \For{$c$ in $C^r$} 
        \State Compute similarity weights $u_c$ using \textbf{Eqs.~\ref{eqn_sim}} and \textbf{\ref{eqn_sim_weights}}
        \State Compute sample weights $v_c$ using \textbf{Eq.~\ref{eqn_sample_weights}}
    \EndFor 
    
    \For{$c$ in $C^r$} 
        \State Compute aggregation weights $w_c$ using \textbf{Eq.~\ref{eqn_fsim_weights}}
    \EndFor  
    
    \If{$r > 10$}
        \State Regularize the aggregation weights $w_c$ using \textbf{Eq.~\ref{eqn_fsim_regularization}}
    \EndIf
    
    \State Compute master model parameters $p^m$ using \textbf{Eq.~\ref{eqn_params}}
    
    \State \textbf{return} $p^m$
\EndProcedure
\end{algorithmic}
\end{algorithm}

\section{Experiments}
\label{sec:results}

\subsection{Setup}
Task 1 focuses on efficient aggregation, client selection, training-per-round, and communication efficiency in order to optimize the federation process. We have devised a mechanism for aggregating model updates trained on individual collaborators that is both efficient and effective. A training data set with total of 1251 multi-institutional patients and 219 validation data set was available. Supplementary data shows how patients are divided into distinct partitions. Partition 1 has 23 contributors, whereas partition 2 has 33 collaborators. For the semantic segmentation of the total tumor, tumor core, and enhancing tumor, the experimental setup leverages Intel's OpenFL platform for federated learning and a preconfigured 3D U-shape neural network. Binary DICE similarity (total tumor, enhancing tumor, tumor core) and Hausdorff (95 percent) distance are the metrics computed in the aggregation rounds (whole tumor, enhancing tumor, tumor core) as described in \cite{pati2021federated}.

The hyperparameters used are shown in Table~\ref{turns_Hyperparameters}. Collaborator selection for RegSimAgg is shown in Fig.~\ref{collaborators}. 

\begin{table}[ht]
\caption{Hyperparameters used in aggregation algorithms.}
\label{turns_Hyperparameters}
\begin{tabular*}{\textwidth}{@{\extracolsep{\fill}} l *{2}l}
\toprule
Hyperparameter & \multicolumn{1}{l}{RegSimAgg}\\
\midrule
Learning rate & 5e-5 \\
Epochs per round & 1.0  \\
\bottomrule
\end{tabular*}
\end{table}

\subsection{Results}

In this section, regularized similarity weighted aggregation (RegSimAgg) findings are summarized for partition 2. The results demonstrate that our approach quickly converges and maintains stability as learning advances across all assessed criteria.

\subsubsection{Model training and performance using internal validation data.}

\begin{figure*}[!th] 
\centering\includegraphics[width=\textwidth]{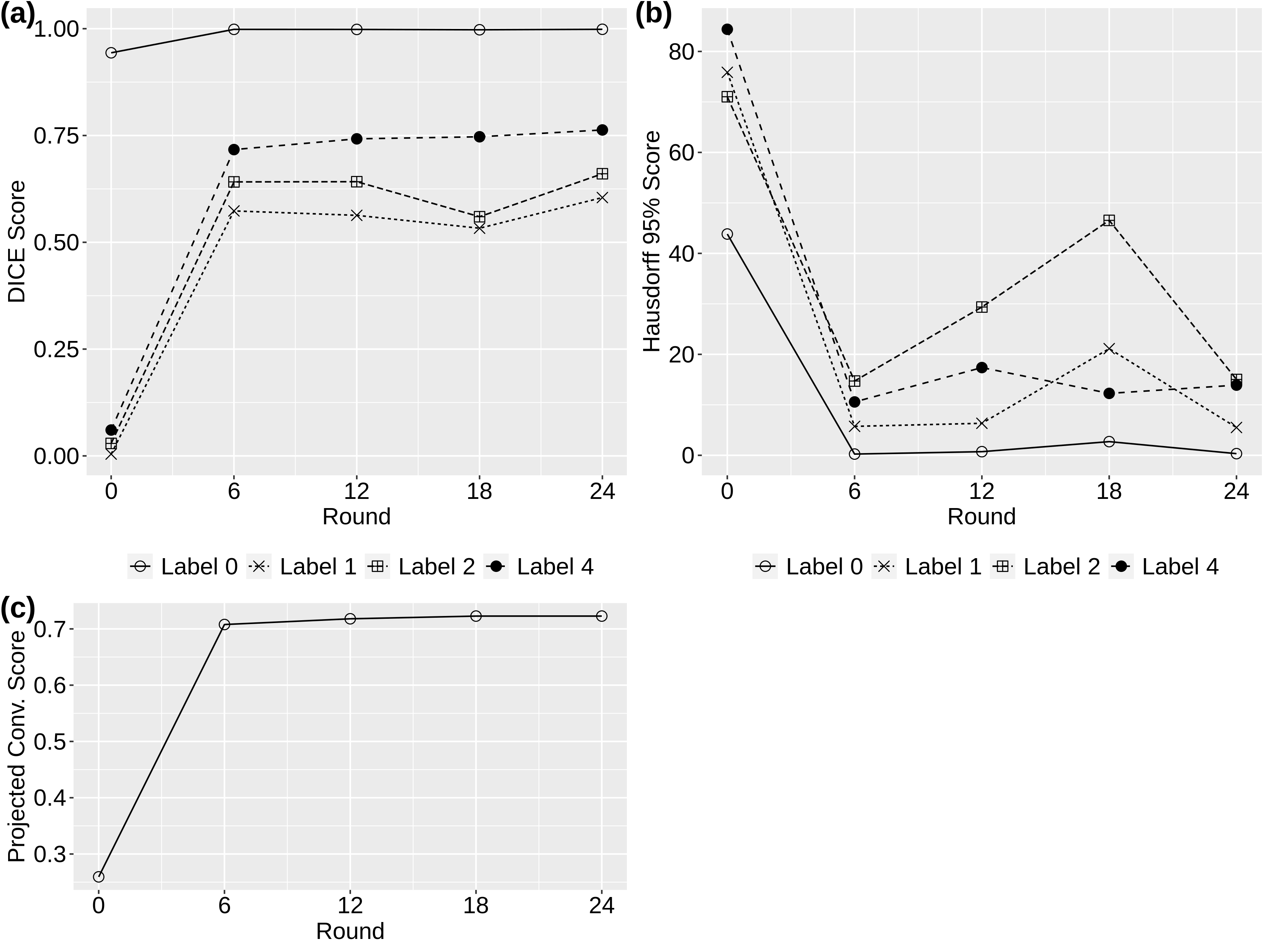}
\caption{Performance metrics for model training of RegSimAgg for partition 2 for internal validation. The horizontal axis refers to the number of rounds and the vertical axis to the performance metrics.
 (a) DICE Score for labels 0, 1, 2, 4; (b) Hausdorff 95\% Score for labels 0, 1, 2, 4; (c) Projected Convergence Score. Total simulation time was approximately 154 hours.}
\label{fig:RegSimAgg_training_p2}
\end{figure*}

Figure~\ref{fig:RegSimAgg_training_p2} shows the performance of model training on internal validation data for partition 2. 



\subsubsection{Model performance using external validation data.}
Prior to the formal testing phase, challenge organizers provided 219 cases of external validation data that we used to evaluate the performance of our approach RegSimAgg, see Tables~\ref{tab_external_validation_p2}. Overall, RegSimAgg performs better on whole tumor segmentation as compared to enhancing tumor segmentation and tumor core segmentation.


\begin{table}[ht]
\caption{Performance on external validation data for partition 2.}
\label{turns}
\begin{tabular*}{\textwidth}{@{\extracolsep{\fill}} l *{2}l}
\toprule

Metrics & \multicolumn{1}{l}{RegSimAgg} \\

\midrule
Dice ET      &  0.7350\\
Dice TC      & 0.7337\\
Dice WT      &  0.8091\\
Hausdorff95 ET      & 30.3497\\
Hausdorff95 TC      & 25.3156\\
Hausdorff95 WT      & 23.7706\\
Sensitivity ET      & 0.7231\\
Sensitivity TC      & 0.7146\\
Sensitivity WT      & 0.8144\\
Specificity ET      & 0.9998\\
Specificity TC      & 0.9998\\
Specificity WT      & 0.9988\\

\bottomrule
\end{tabular*}
\label{tab_external_validation_p2}
\end{table}

\begin{table}[ht]
\setlength\tabcolsep{0pt} 
\caption{RegSimAgg (HT-TUAS) test set performance on Leaderboard}
\begin{tabular*}{\textwidth}{@{\extracolsep{\fill}} l *{5}c}
\toprule
 & \multicolumn{1}{c}{Mean} & \multicolumn{1}{c}{Standard Deviation} & \multicolumn{1}{c}{Median} & \multicolumn{1}{c}{25 quantile} & \multicolumn{1}{c}{75 quantile} \\
\midrule
DICE ET      & 0.6745 & 0.2920 & 0.8004 & 0.5704 & 0.8866   \\

DICE WT      & 0.7247 & 0.2026 & 0.7820 & 0.6235 & 0.8811   \\

DICE TC      & 0.7169 & 0.3002 & 0.8655 & 0.6154 & 0.9259   \\

Sensitivity ET      & 0.7501 & 0.3087 & 0.8815 & 0.6882 & 0.9612   \\
 
Sensitivity WT      & 0.7834 & 0.2044 & 0.8546 & 0.7079 & 0.9307   \\
              
Sensitivity TC 
              & 0.7592 & 0.3091 & 0.9035 & 0.6990 & 0.9715   \\

Specificity ET      & 0.9991 & 0.0017 & 0.9997 & 0.9992 & 0.9999   \\

Specificity WT      & 0.9974 & 0.0033 & 0.9985 & 0.9968 & 0.9994   \\

Specificity TC      & 0.9992 & 0.0017 & 0.9997 & 0.9994 & 0.9999   \\

Hausdorff (95\%) ET      & 35.2283 & 86.9517 & 3.7417 & 1.7321 & 19.1724   \\

Hausdorff (95\%) WT      & 35.9036 & 30.8538 & 31.6172 & 8.4853 & 58.6204   \\

Hausdorff (95\%) TC 
              & 33.7853 & 80.5775 & 6.4807 & 3.0000 & 20.1239   \\

Communication Cost 
              & 0.6905 & 0.6905 & 0.6905 & 0.6905 & 0.6905   \\

\bottomrule
\end{tabular*}
\label{tab_test_LB1}
\end{table}


\subsubsection{Model performance using fully blinded test set.}

Team HT-TUAS submitted RegSimAgg algorithm for the leaderboard ranking. Model training is performed for 500 rounds by challenge organizers. The performance stats on the fully blinded test set for 570 patients are shown in Tables~\ref{tab_test_LB1}. When a weightage of 6 is given to the communication cost, our team achieved third place in the leaderboard. 





\section{Discussion}
\label{sec:discussion}

Several federated aggregation methods like exponential smoothing aggregation and conditional threshold aggregation methods require user defined threshold parameter settings. Therefore, these algorithms are not easily applicable to new and unexplored data sets. To overcome this issue, we developed regularized similarity weighted aggregation, which adaptively learns the participating collaborator weights.

Our method does not require any participation in the modeling process at the client side but it learns a global model at the server side. It is able to minimize the contribution of diverging collaborators and allows clients with varying settings to join the federation. This is in contrast to Fedprox~\cite{li2018federated}, which performs client side regularization of diverging collaborators. Moreover, studies have shown that training federated learning algorithms using a randomly selected subset of collaborators expedites the process~\cite{zhao2018federated}. We expanded on these ideas and developed a sliding window technique that corroborates that all collaborators are participate in the training process.

FeTS 2022 data is released in two partitions which split the training data into non-IID data sets based on institution and tumor size. As a result, amount and characteristics of data may vary for different collaborators. Weighted aggregation strategy helps to learn a model that represents well the majority of the contributors at each round with low impact by outliers. Therefore, our method works well on both partitions. Even though the model works well when data has non-IID splits in general, it could be also be useful to study the performance on outliers further.

A limitation of the current setting is that the number of patients for each collaborator is fixed. In a real-world setting the patient data at a collaborator may change between federation rounds. Additional patient data potentially alters collaborator's data distribution -- which might, in turn, affect the learned parameters at the master model: during the FL learning process, a previously diverging collaborator may become similar to the participating collaborators, compelling some other participating collaborators to become outliers. 
However, our approach takes diverging collaborator weights into account at each round. Hence, this method can be used as a baseline for refined generation of a better model that can be used at scale to newly generated data sets. Moreover, incorporation of our FL approach with a cutting-edge privacy protection AI framework is one of our future research directions. We also intend to investigate a potential solution to the communication payload bottleneck between the collaborators and the server. Further, we will also study how a decrease in the payload throughput affects model convergence time-efficiency and task performance optimization.

\section{Conclusion}
\label{sec:conclusion}

In this paper, we propose regularized similarity weighted aggregation as an improved weight aggregation strategy for federated learning and apply it to imaging in order to achieve robust brain tumor segmentation. In our experiments, the proposed method -- deployed in the OpenFL platform -- performed well in terms of convergence score and communication costs. Currently, the proposed algorithm is a comprehensive proof-of-concept (POC) solution in the healthcare domain, demonstrated to potentially assist radiologists in diagnostic digital pathology. However, this edge computing infrastructure can easily be scaled to versatile real-world multi-client production level applications for foundational collaborative computation and federated learning workflows in other disciplines like internet-of-things (IoT) and telecommunication. Moreover, this methodology can issue stronger privacy guarantee via integrating differential privacy or secure multi-party computation, or a combination of the two, which is an intriguing future research topic.

\section{Acknowledgements}
This work was supported by the Business Finland under Grant 33961/31/2020. We also acknowledge the support and computational resources facilitated by the CSC-Puhti super-computer, a non-profit state enterprise owned by the Finnish state and higher education institutions in Finland.

%
%
%
 \bibliographystyle{splncs04}
 \bibliography{main.bib}

\end{document}